\documentclass[11pt]{article}

\usepackage[preprint]{automl}

\usepackage[utf8]{inputenc}
\usepackage[T1]{fontenc}
\usepackage{natbib}
\usepackage{hyperref}
\usepackage{url}
\usepackage{booktabs}
\usepackage{amsfonts}
\usepackage{amsmath}
\usepackage{nicefrac}
\usepackage{microtype}
\usepackage{xcolor}
\usepackage{graphicx}
\usepackage{subcaption}
\usepackage{multirow}

\graphicspath{{figures/}}

\title{AGOP-IxG: A Gradient Covariance Filter\\
for Local Feature Attribution on Tabular Data,\\
with a Controlled Benchmark}

\author[1]{\nameemail{Raj Kiran Gupta Katakam}{raj.katakam@creditkarma.com}}
\affil[1]{Credit Karma}

\begin{document}

\maketitle

\begin{abstract}
Automated machine learning pipelines increasingly produce models whose
predictions must be explained to end users, auditors, and downstream
decision systems. The most widely used feature attribution methods
(SHAP, Integrated Gradients, LIME) are typically chosen by convention
rather than measured fidelity, because rigorous evaluation is impeded
by the absence of ground-truth attribution on real data. We propose
\emph{AGOP-IxG}, a fast per-sample attribution method for tabular
classifiers that pre-multiplies the per-sample gradient by a top-$K$
rank-truncated Average Gradient Outer Product matrix, and evaluate
it against four widely-used baselines on a controlled tabular
benchmark designed for AutoML practitioners.
In Part~1, we construct three synthetic multi-class tabular tasks
--- linear, sparse nonlinear, and interaction-based --- where
ground-truth attribution per sample is analytically or numerically
derivable, and compare five methods: AGOP-IxG (shortened to
``AGOP'' in figure panels only), SHAP (DeepExplainer), Integrated
Gradients, InputXGradient, and LIME.
AGOP-IxG leads on Spearman rank correlation and on noise feature mass
on \emph{all three} synthetic datasets, and on top-$k$ precision on
the interaction dataset. On the linear task, LIME has a marginal
advantage on top-$k$ precision only ($0.915$ vs.\ $0.895$). On the
sparse nonlinear task, SHAP holds a near-tie advantage on top-$k$
precision ($0.515$ vs.\ AGOP-IxG's $0.514$) but AGOP-IxG leads on
Spearman ($0.516$ vs.\ $0.511$) and on noise feature mass by a wide
margin ($0.244$ vs.\ $0.409$). Across all settings, AGOP-IxG is
approximately $350\times$ to $1{,}650\times$ faster than SHAP
($\sim$0.02--0.10\,s vs.\ $\sim$33\,s per dataset). In Part~2, we evaluate global faithfulness on two real
tabular datasets (Adult Income and Credit Card Default) using the
ROAR protocol \citep{hooker2019benchmark}. On real data the methods
are closely clustered, and no single method dominates, consistent
with AGOP-IxG being optimized for per-sample local attribution rather
than global feature ranking.
\end{abstract}

\section{Introduction}
\label{sec:intro}

Modern AutoML systems search over models, hyperparameters, and
preprocessing pipelines, with the result that the model handed to a
practitioner is often a black-box artifact of the search rather than
a hand-designed architecture. In high-stakes deployments --- credit
scoring, medical risk, fraud detection --- regulators and downstream
consumers of the model demand per-prediction explanations. Feature
attribution methods are the standard tooling for this need: they
assign a scalar importance score to each input feature, indicating
its contribution to a specific prediction.

Despite a rich literature, a fundamental evaluation problem remains:
\emph{how do we know if an attribution is correct?} On real tabular
data, ground-truth attribution is unavailable. The most commonly used
proxy --- agreement between methods, or plausibility judged by domain
experts --- is circular: it rewards methods that agree with each
other, not methods that recover the true causal signal. The ROAR
protocol \citep{hooker2019benchmark} offers a model-based proxy by
measuring accuracy degradation under feature removal and retraining,
but it measures global feature importance, not the quality of
individual per-sample explanations. As a consequence, AutoML
practitioners selecting an attribution method for a downstream
pipeline have no principled basis for comparison.

In this paper we propose \emph{AGOP-IxG} (Section~\ref{sec:agop}), a
fast per-sample attribution method for tabular classifiers, and
evaluate it on a controlled benchmark with known ground-truth
attribution per sample, enabling direct evaluation of per-sample
attribution quality without proxy measures. We compare AGOP-IxG
against four widely-used baselines spanning the main algorithmic
families --- gradient-based (InputXGradient), path-integral
(Integrated Gradients), game-theoretic (SHAP), and local surrogate
(LIME) --- on these synthetic benchmarks, and then apply ROAR on two
real datasets to assess global faithfulness.

\paragraph{Relevance to AutoML.}
The benchmark is designed for direct reuse in AutoML evaluation
pipelines: (i)~the synthetic generators are parametric and require no
proprietary data, (ii)~the evaluation metrics are model-agnostic and
applicable to any attribution method that produces a per-sample
attribution vector, and (iii)~ground-truth attribution provides a
loss-like signal that can be used inside an AutoML system to
auto-select or auto-tune attribution methods for a target model.

Our main contributions are:
\begin{enumerate}
  \item \emph{AGOP-IxG} (Section~\ref{sec:agop}): a fast per-sample
    attribution method for tabular classifiers composing
    InputXGradient with a top-$K$ rank-truncated AGOP gradient
    filter. AGOP-IxG achieves the lowest noise-feature mass on every
    synthetic benchmark we evaluate.
  \item A controlled tabular evaluation framework
    (Section~\ref{sec:synthetic}) with three multi-class synthetic
    regimes (linear, sparse nonlinear, feature-interaction) for which
    per-sample ground-truth attribution is analytically or
    numerically derivable, evaluated under Spearman, top-$k$, and
    noise feature mass across four seeds.
  \item A ROAR evaluation on real tabular data
    (Section~\ref{sec:roar}) contrasting local per-sample fidelity
    (where AGOP-IxG leads) against global feature ranking (where
    methods cluster within ${\sim}1.7\%$ AUC) --- a distinction
    relevant when AutoML systems must serve both per-decision
    explanations and feature-selection consumers.
\end{enumerate}

\section{Background and Related Work}
\label{sec:related}

\paragraph{Gradient-based attribution.}
\citet{simonyan2014deep} introduced vanilla gradients as saliency
maps. \citet{shrikumar2016not} showed that gradients times inputs
better reflect feature contributions, leading to InputXGradient.
\citet{sundararajan2017axiomatic} formalised Integrated Gradients
(IG), which satisfies the completeness axiom by integrating gradients
along a path from a baseline to the input.

\paragraph{Game-theoretic attribution.}
\citet{shapley1953value} defined Shapley values in cooperative game
theory; \citet{lundberg2017unified} unified several attribution
methods under the SHAP framework and provided DeepExplainer for
neural networks via a background-dataset approximation.

\paragraph{Local surrogate methods.}
\citet{ribeiro2016should} proposed LIME, which fits a local linear
model in the neighbourhood of each sample using perturbed inputs.

\paragraph{Second-order gradient methods.}
The Average Gradient Outer Product (AGOP) matrix $M = \frac{1}{n}
\sum_i g_i g_i^\top$ has connections to the Neural Tangent Kernel
\citep{jacot2018neural} and to function-space geometry studied in
\citet{radhakrishnan2022mechanism}. \citet{engstrom2023dsdm} use
gradient covariance for data selection; our work applies it as a
per-sample attribution filter. \citet{radhakrishnan2022mechanism}
studies AGOP in the context of feature learning in neural networks
but does not apply it as an attribution method or benchmark it against
SHAP/IG/LIME on controlled ground-truth tasks. Concurrent
work~\citep{katakam2026agop} explores AGOP-based attribution for
\emph{image} classifiers via a different formulation
(\emph{AGOP-Weighted}: per-pixel scaling of the gradient magnitude by
the normalised diagonal $\sqrt{\mathrm{diag}(M)/\max_j
\mathrm{diag}(M)_j}$). The AGOP-IxG formulation in this paper
(Section~\ref{sec:agop}) instead uses the full top-$K$ rank
reconstruction $M_K = V_K\Lambda_K V_K^\top$ as a linear projection
of the gradient, with InputXGradient-style baseline subtraction; the
two methods share the AGOP matrix as a training-distribution prior
but apply it differently.

\paragraph{Evaluation of attribution methods.}
\citet{hooker2019benchmark} proposed ROAR, replacing top-attributed
features and retraining; we use this in Part~2.
\citet{samek2017evaluating} proposed pixel-flipping for images.
\citet{adebayo2018sanity} introduced sanity checks via model/data
randomisation. Synthetic benchmarks with known ground truth have
been used for image saliency \citep{yang2019benchmarking} but are
less common for tabular data; our work fills this gap.

\paragraph{Attribution in AutoML.}
Attribution outputs are increasingly produced by AutoML systems
alongside the model itself (e.g.\ as part of a deployed
explainability service). Most AutoML benchmarks
\citep{gijsbers2022amlb} measure predictive accuracy and runtime, not
the quality of the explanations the system produces. Our benchmark
provides the missing complementary measurement.

\section{Methods}
\label{sec:methods}

\subsection{AGOP-IxG Attribution}
\label{sec:agop}

The AGOP matrix \citep{radhakrishnan2022mechanism} is a definition,
not an attribution method. We propose \emph{AGOP-IxG}, an attribution
scheme that composes InputXGradient with a top-$K$ AGOP gradient
filter. We use ``AGOP-IxG'' throughout the body and tables to refer
to the method; the in-figure panel titles use the short label
``AGOP'' for legibility.

\paragraph{Stage 1 --- Fit (training data).}
For each training sample $x_i$, compute the gradient of the predicted
class score:
\begin{equation}
  g_i = \nabla_x\, \hat{s}_{c_i}(x_i),
  \quad c_i = \arg\max_c\, f_c(x_i).
\end{equation}
The AGOP matrix is the empirical gradient outer-product (a positive
semi-definite gradient ``covariance'' about the origin):
\begin{equation}
  M = \frac{1}{n} \sum_{i=1}^{n} g_i g_i^\top \;\in\; \mathbb{R}^{d \times d}.
\end{equation}
Eigendecompose $M = V\Lambda V^\top$ and retain the top-$K$ eigenvectors
whose eigenvalue exceeds $1\%$ of $\lambda_{\max}$.
Define the factor $M_{\sqrt{},K} = V_K \cdot
\mathrm{diag}(\sqrt{\lambda_1},\ldots,\sqrt{\lambda_K})
\in \mathbb{R}^{d \times K}$, so that
$M_{\sqrt{},K}\, M_{\sqrt{},K}^\top = V_K \Lambda_K V_K^\top = M_K$,
where $M_K$ denotes the rank-$K$ reconstruction of $M$.

\paragraph{Stage 2 --- Attribute (test sample $x$, baseline $x'$).}
\begin{align}
  g &= \nabla_x\, \hat{s}_c(x), \quad c = \arg\max_c f_c(x), \\
  \alpha &= g \cdot M_{\sqrt{},K} \in \mathbb{R}^K, \\
  g_\mathrm{attr} &= \alpha \cdot M_{\sqrt{},K}^\top \in \mathbb{R}^d, \\
  e_j &= (x_j - x'_j)\cdot g_{\mathrm{attr},j}, \\
  r_j &= |e_j| \,/\, \textstyle\sum_k |e_k|.
\end{align}
The baseline $x'$ is the per-feature training mean.

\paragraph{Simplification and interpretation.}
Combining the two factor-multiplications collapses Stage 2 to a single
linear transformation:
\begin{equation}
  g_\mathrm{attr}
   = g\,M_{\sqrt{},K}\,M_{\sqrt{},K}^\top
   = g\,M_K,
   \qquad
  e_j = (x_j - x'_j)\,(g\,M_K)_j.
  \label{eq:agop_simplification}
\end{equation}
AGOP-IxG is therefore exactly InputXGradient applied to a gradient
pre-multiplied by the truncated AGOP matrix $M_K$.

\paragraph{Why this prior?}
$M$ is the empirical second-moment of training-set gradients; its
dominant eigendirections capture consistent score variation, while
low-eigenvalue directions are dominated by per-sample noise.
Pre-multiplying by $M_K$ \emph{amplifies} gradient components aligned
with high-variance training-distribution directions and zeros out
low-eigenvalue noise directions. This differs both from
natural-gradient preconditioning by $M^{-1}$ (which \emph{whitens}
all directions, equalising signal and noise) and from the
diagonal-only AGOP-Weighted form for image classifiers
\citep{katakam2026agop} which discards the off-diagonal mixing the
full $M_K$ provides. Two further properties make this design
operationally attractive for AutoML pipelines:
\begin{itemize}
  \item \emph{Explicit noise truncation.} The threshold
    $\lambda_k > 0.01\,\lambda_{\max}$ zeros out eigendirections
    dominated by noise; on the linear, sparse, and interaction
    datasets the surviving rank is $K{=}2,12,19$ respectively
    (Appendix~\ref{app:impl}). The low ``noise feature mass'' in
    Table~\ref{tab:synthetic} is the empirical signature of this
    truncation.
  \item \emph{Low-rank compute.} For $K\!\ll\!d$ we never form the
    $d{\times}d$ matrix $M_K$; we keep only the $d{\times}K$ factor
    $M_{\sqrt{},K}$ and apply it via two matrix--vector products.
\end{itemize}

\subsection{Baseline Methods}

\textbf{InputXGradient} \citep{shrikumar2016not}:
$e_j = x_j\cdot \partial \hat{s}_c / \partial x_j$ with $c = \arg\max_c f_c(x)$,
matching \texttt{captum.attr.InputXGradient}. No fitting step.

\textbf{Integrated Gradients} \citep{sundararajan2017axiomatic}:
$e_j = (x_j - x'_j)\int_0^1 \partial\hat{s}_c/\partial x_j(x' +
\alpha(x-x'))\,d\alpha$, approximated with 50 right-endpoint Riemann
steps. The target class $c = \arg\max_c f_c(x)$ is pinned to the
prediction at $x$ throughout the integration path, so the completeness
axiom holds.

\textbf{SHAP} \citep{lundberg2017unified}: DeepExplainer with 200
background training samples. For each sample, we use the SHAP values
for the predicted class only: $e_j = |\phi_{j,c}|$ with $c = \arg\max_c f_c(x)$.
For binary classification (Part~2), this is equivalent up to a constant
factor to the alternative aggregation $\sum_c |\phi_{j,c}|$ because
completeness implies $|\phi_{j,0}| \approx |\phi_{j,1}|$.

\textbf{LIME} \citep{ribeiro2016should}: \texttt{LimeTabularExplainer}
with 1000 neighbourhood samples, continuous features, predicted-class
output. For ROAR global ranking, 500 training samples are used due to
per-sample fitting cost.

\paragraph{Normalization.}
All methods produce $r_j = |e_j| / (\sum_k |e_k| + \epsilon)$,
$\epsilon{=}10^{-8}$, so attributions are non-negative and sum to~1.

\section{Part 1: Synthetic Benchmarks}
\label{sec:synthetic}

\subsection{Model}
A single MLP per dataset: Input $\to$ 256 $\to$ 256 $\to$ 128 $\to$
64 $\to$ $C$ (ReLU activations, no BatchNorm, no Dropout).
Adam, lr$=10^{-3}$, 200 epochs, batch 256, CUDA GPU.
The same saved model is used for all attribution methods.

\subsection{Datasets}
All datasets: $n{=}5000$, $d{=}20$, $C{=}3$,
80/20 stratified split. Features standardised; baseline $x'$ =
training mean.
Results in Section~\ref{sec:results} are mean $\pm$ std over four
independent seeds (0, 1, 2, 42), each generating a separate
dataset, model, and attribution run.

\paragraph{Linear.}
$W \in \mathbb{R}^{3\times 5}$, entries $\sim \mathrm{Uniform}(0.5,2.0)$
with random signs.
$y = \arg\max(X_{:5}W^\top + \epsilon)$, $\epsilon \sim
\mathcal{N}(0,0.3^2)$. Features $f_5,\ldots,f_{19}$ are pure noise.
Ground truth: $\mathrm{attr}_j = |W_{c,j}(x_j{-}\bar{x}_j)| /
\sum_{l<5}|W_{c,l}(x_l{-}\bar{x}_l)|$ (closed form).
Test accuracy: \textbf{92.1\%}.

\paragraph{Sparse nonlinear.}
$s_0{=}\sin(x_0)+x_1^2$, $s_1{=}x_2 x_3+x_4$,
$s_2{=}\cos(x_0{+}x_1)+x_3^2$; $y{=}\arg\max(s+\epsilon)$.
Features $f_5,\ldots,f_{19}$ are noise.
Ground truth via central finite differences on the true score function
($\epsilon{=}10^{-4}$), then input$\times$gradient and normalise.
Test accuracy: \textbf{85.1\%} (Bayes floor from label noise).

\paragraph{Interaction.}
$s_0{=}x_0 x_1+x_2$, $s_1{=}x_3 x_4+x_5$,
$s_2{=}x_0 x_3+x_1 x_4$; $y{=}\arg\max(s+\epsilon)$.
Informative features: $f_0,\ldots,f_5$ ($k{=}6$).
Ground truth via same finite-difference approach.
Test accuracy: \textbf{84.0\%}.

\subsection{Evaluation Metrics}

\textbf{Spearman rank correlation} $\rho$: mean per-sample Spearman
$\rho_s(\mathrm{true\_attr}_i, \mathrm{pred\_attr}_i)$.
Higher is better; range $[-1,1]$.

\textbf{Top-$k$ precision}: $\frac{1}{N}\sum_i
|\mathcal{T}_i \cap \mathcal{P}_i|/k$, where $\mathcal{T}_i,
\mathcal{P}_i$ are ground-truth and predicted top-$k$ feature sets.
Higher is better.

\textbf{Noise feature mass}: mean fraction of attribution assigned to
noise features (true attr $=0$). Lower is better.

\paragraph{Evaluation subset.}
All three metrics are computed only on \emph{correctly-classified} test
samples (model prediction $= y_{\text{test}}$).
Misclassified samples are excluded because the model may use different
features to arrive at a wrong answer, and ground-truth attribution
(derived from the noise-free score function) would measure a quantity inconsistent
with the model's actual computation.

\subsection{Results}
\label{sec:results}

\begin{table}[t]
\centering
\caption{Attribution quality on synthetic benchmarks (mean $\pm$ std, 4 seeds: 0, 1, 2, 42).
  All methods evaluated on the correctly-classified subset of the
  1{,}000-sample test set (Linear $\approx$921, Sparse $\approx$851,
  Interaction $\approx$840 samples per seed; see Section~\ref{sec:results}, ``Evaluation subset'').
  \textbf{Bold} = best per dataset per metric.}
\label{tab:synthetic}
\footnotesize
\setlength{\tabcolsep}{4pt}
\begin{tabular}{llcccr}
\toprule
Dataset & Method & Spearman $\uparrow$ & Top-$k$ Prec $\uparrow$ & Noise Mass $\downarrow$ & Time (s) \\
\midrule
\multirow{5}{*}{Linear}
  & \textbf{AGOP-IxG}  & $\mathbf{0.707\pm.027}$ & $0.895\pm.040$ & $\mathbf{0.066\pm.005}$ & 0.095 \\
  & LIME           & $0.696\pm.041$ & $\mathbf{0.915\pm.051}$ & $0.172\pm.009$ & 15.04 \\
  & SHAP           & $0.642\pm.047$ & $0.801\pm.055$ & $0.160\pm.019$ & 32.98 \\
  & IntGrad        & $0.559\pm.061$ & $0.685\pm.067$ & $0.245\pm.037$ & 1.98 \\
  & InputXGrad     & $0.521\pm.048$ & $0.643\pm.051$ & $0.299\pm.027$ & \textbf{0.01} \\
\midrule
\multirow{5}{*}{Sparse}
  & \textbf{AGOP-IxG} & $\mathbf{0.516\pm.005}$ & $0.514\pm.012$ & $\mathbf{0.244\pm.005}$ & 0.02 \\
  & SHAP           & $0.511\pm.004$ & $\mathbf{0.515\pm.004}$ & $0.409\pm.016$ & 32.70 \\
  & IntGrad        & $0.403\pm.008$ & $0.441\pm.005$ & $0.453\pm.009$ & 1.97 \\
  & InputXGrad     & $0.400\pm.006$ & $0.440\pm.004$ & $0.461\pm.011$ & \textbf{0.01} \\
  & LIME           & $0.358\pm.008$ & $0.423\pm.014$ & $0.619\pm.004$ & 14.71 \\
\midrule
\multirow{5}{*}{Interaction}
  & \textbf{AGOP-IxG}  & $\mathbf{0.534\pm.004}$ & $\mathbf{0.521\pm.003}$ & $\mathbf{0.402\pm.005}$ & 0.02 \\
  & SHAP           & $0.466\pm.004$ & $0.478\pm.007$ & $0.496\pm.003$ & 32.64 \\
  & IntGrad        & $0.432\pm.003$ & $0.463\pm.009$ & $0.526\pm.006$ & 1.97 \\
  & InputXGrad     & $0.431\pm.003$ & $0.467\pm.003$ & $0.536\pm.007$ & \textbf{0.01} \\
  & LIME           & $0.230\pm.010$ & $0.367\pm.004$ & $0.629\pm.004$ & 14.40 \\
\bottomrule
\end{tabular}
\end{table}

Table~\ref{tab:synthetic} shows the main results (mean $\pm$ std, 4 seeds).
AGOP-IxG achieves the lowest noise feature mass on every dataset and
leads on Spearman rank correlation on every dataset. The single
metric on which other methods edge AGOP-IxG is top-$k$ precision
(LIME on the linear task; SHAP on sparse by a $0.001$ margin).
On the linear task, AGOP-IxG leads LIME on Spearman ($0.707\pm.027$
vs.\ $0.696\pm.041$), with LIME leading on top-$k$ precision
($0.915$ vs.\ $0.895$) but AGOP-IxG leading substantially on noise mass
($0.066$ vs.\ $0.172$).
LIME's competitiveness on linear data is expected --- a local linear
approximation is well-suited to a globally linear generating process ---
but it collapses on the nonlinear datasets (Spearman $0.358$ on sparse,
$0.230$ on interaction).
On the sparse dataset, AGOP-IxG and SHAP are near-tied on Spearman
($0.516$ vs.\ $0.511$) and on top-$k$ precision ($0.514$ vs.\ $0.515$,
a single-thousandth gap), but AGOP-IxG leads by a wide margin on noise
mass ($0.244$ vs.\ $0.409$).
On the interaction dataset, AGOP-IxG leads SHAP on all three metrics
(Spearman $0.534$ vs.\ $0.466$, a gap of $0.068$ that is over $15\times$
the within-method std), with consistently small standard deviations
($\leq.010$) confirming stable rankings across seeds.
Both nonlinear tasks are accompanied by a large runtime advantage:
AGOP-IxG runs in 0.02--0.095\,s vs.\ $\sim$33\,s for SHAP
(approximately 350--1{,}650$\times$ faster).

\begin{table}[t]
\centering
\caption{Model accuracy and confidence on the test set (canonical seed-42 run).
  Confidence = max softmax probability.}
\label{tab:confidence}
\small
\begin{tabular}{lcccc}
\toprule
Dataset & Test Acc & Mean Conf & Conf (Correct) & Conf (Wrong) \\
\midrule
Linear      & 92.1\% & 0.988 & 0.992 & 0.935 \\
Sparse      & 85.1\% & 0.984 & 0.992 & 0.939 \\
Interaction & 84.0\% & 0.983 & 0.990 & 0.946 \\
\bottomrule
\end{tabular}
\end{table}

Table~\ref{tab:confidence} shows model accuracy and confidence.
The model is highly confident even on wrong predictions ($0.935$--$0.946$),
typical of uncalibrated MLPs trained with cross-entropy loss.
The sparse and interaction accuracy ceiling ($\approx$84\%) reflects
Bayes error from $\sigma{=}0.3$ label noise: the model achieves 100\%
training accuracy, so the gap is irreducible and not a model capacity
issue.

\begin{figure}[t]
  \centering
  \includegraphics[width=0.325\linewidth]{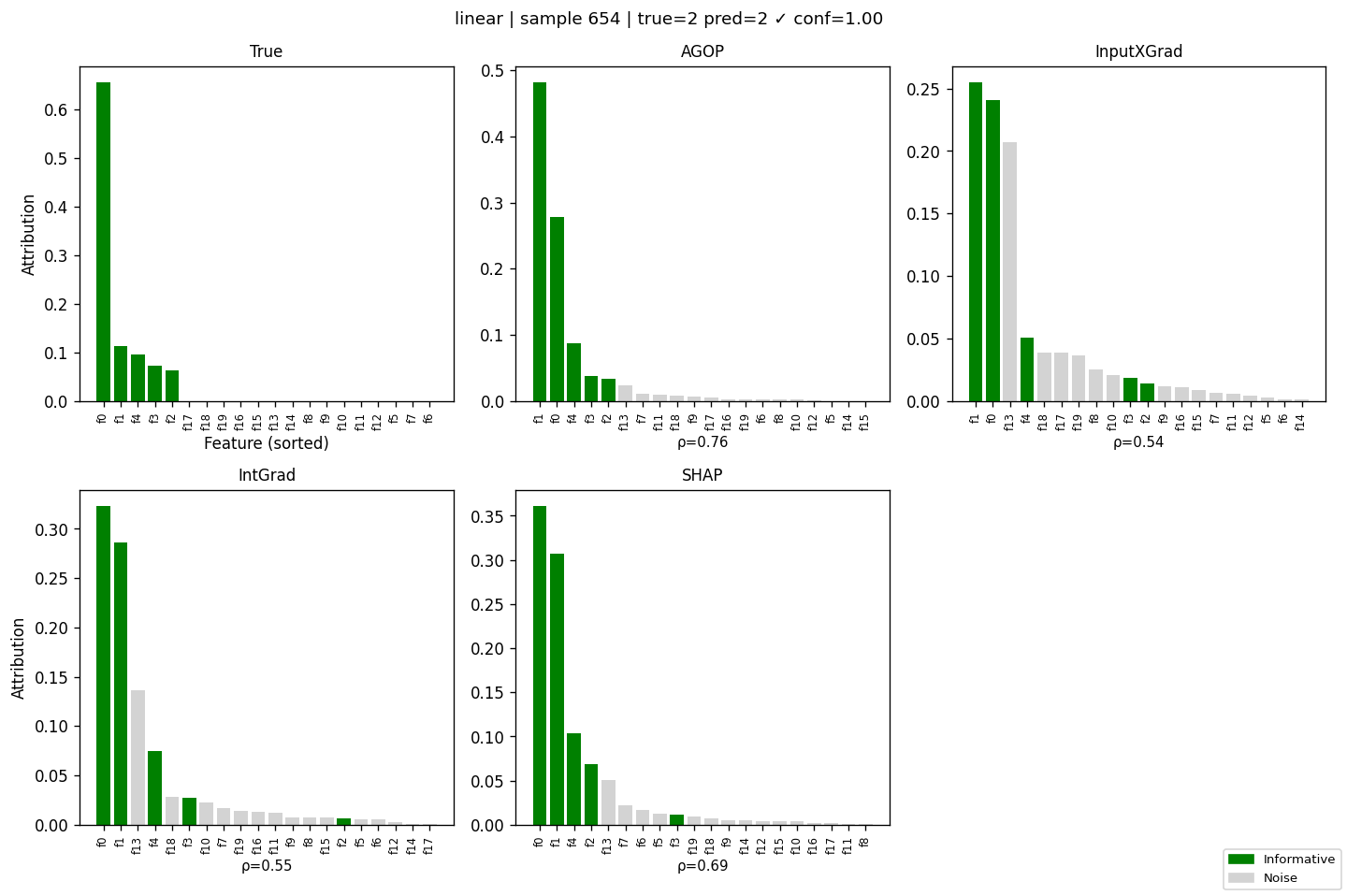}\hfill
  \includegraphics[width=0.325\linewidth]{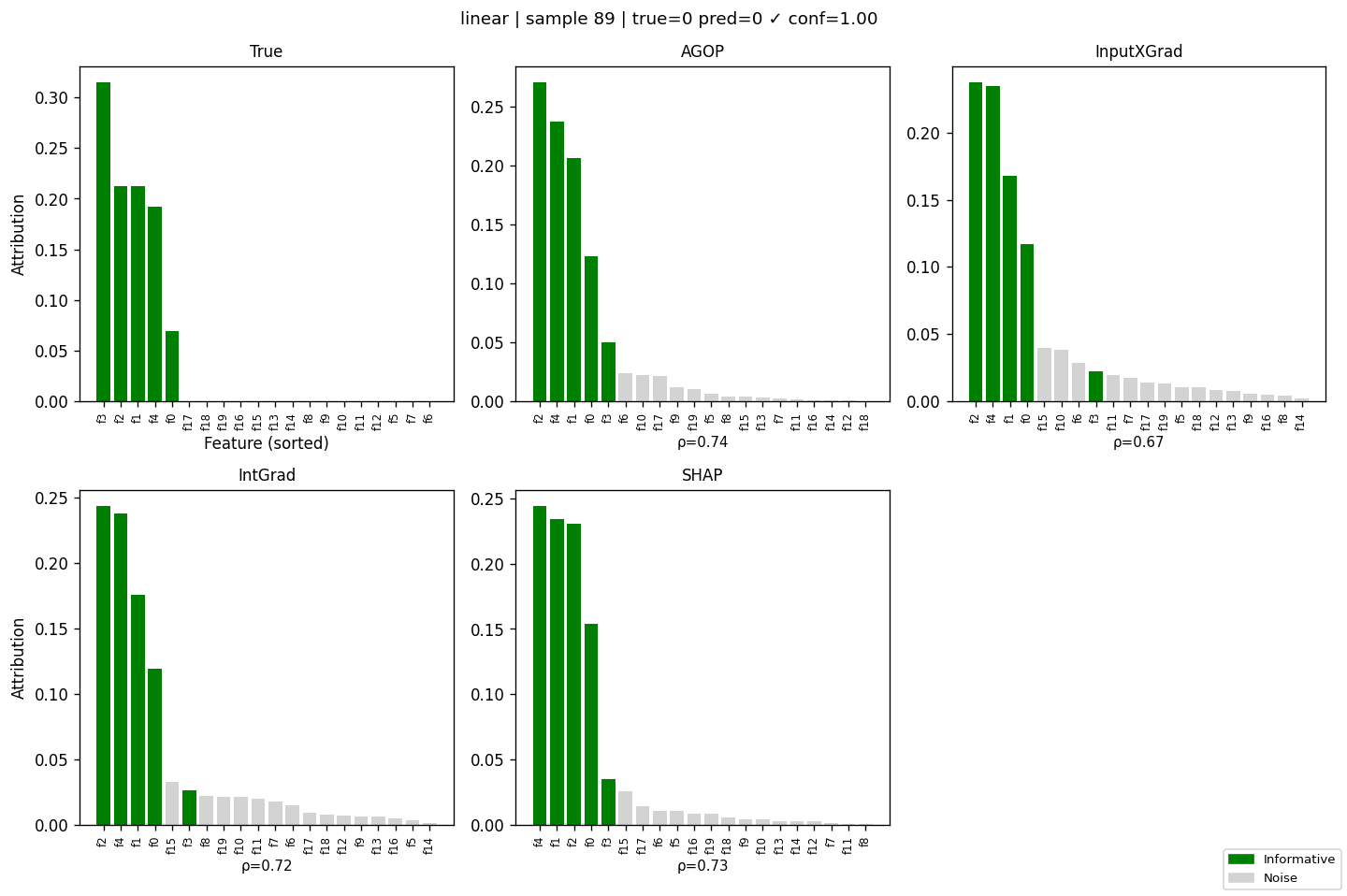}\hfill
  \includegraphics[width=0.325\linewidth]{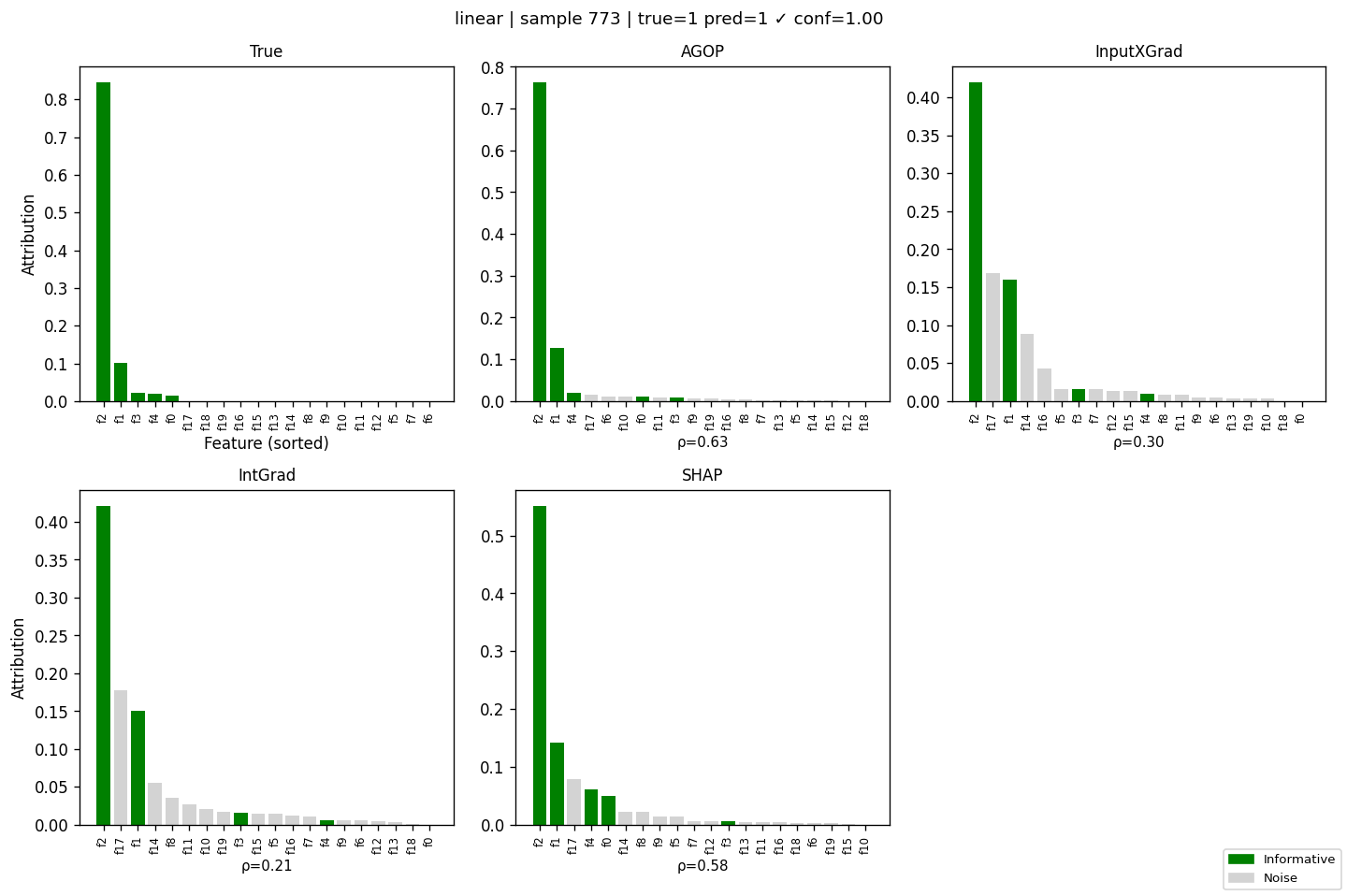}\\[2pt]
  \includegraphics[width=0.325\linewidth]{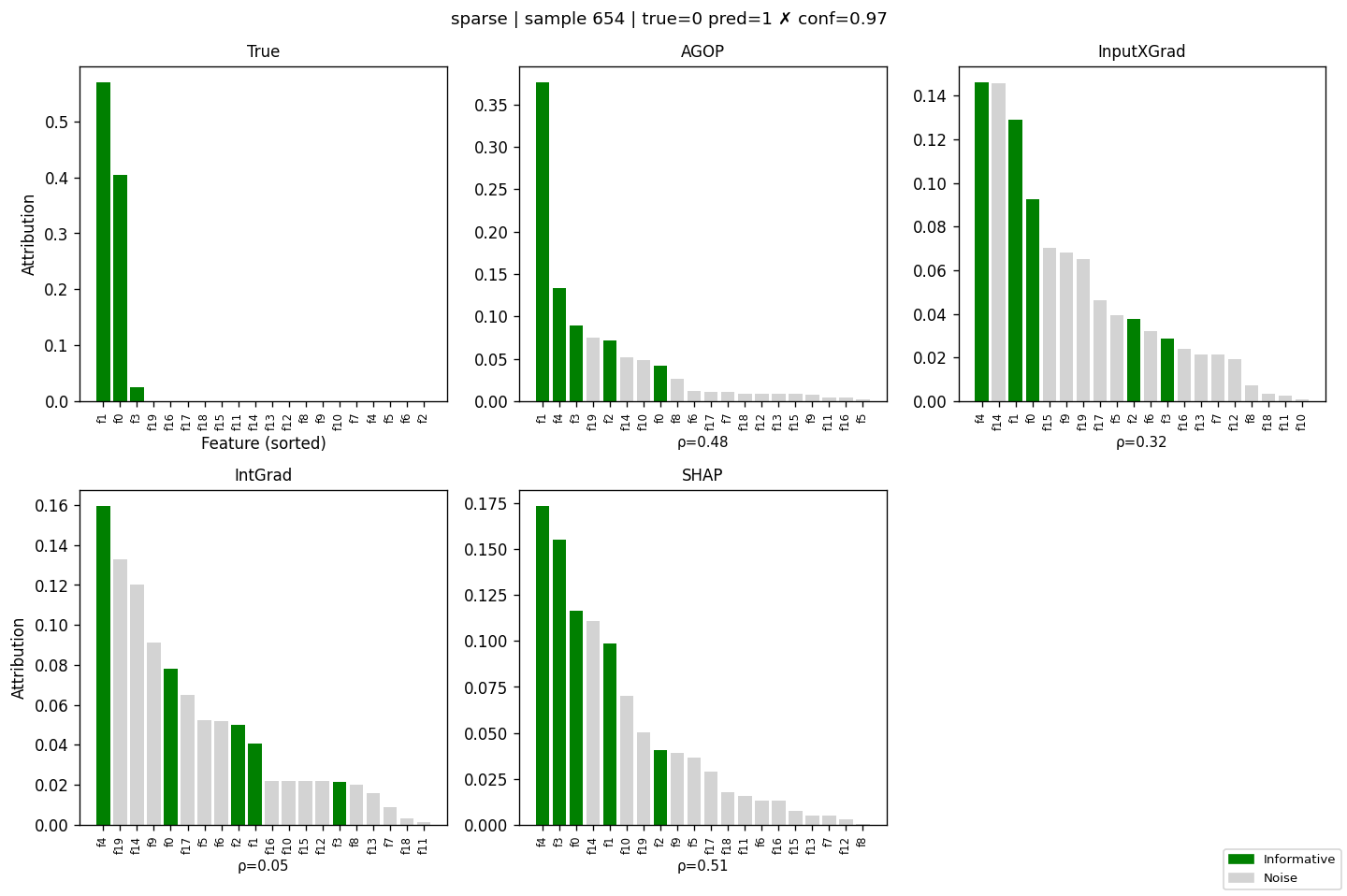}\hfill
  \includegraphics[width=0.325\linewidth]{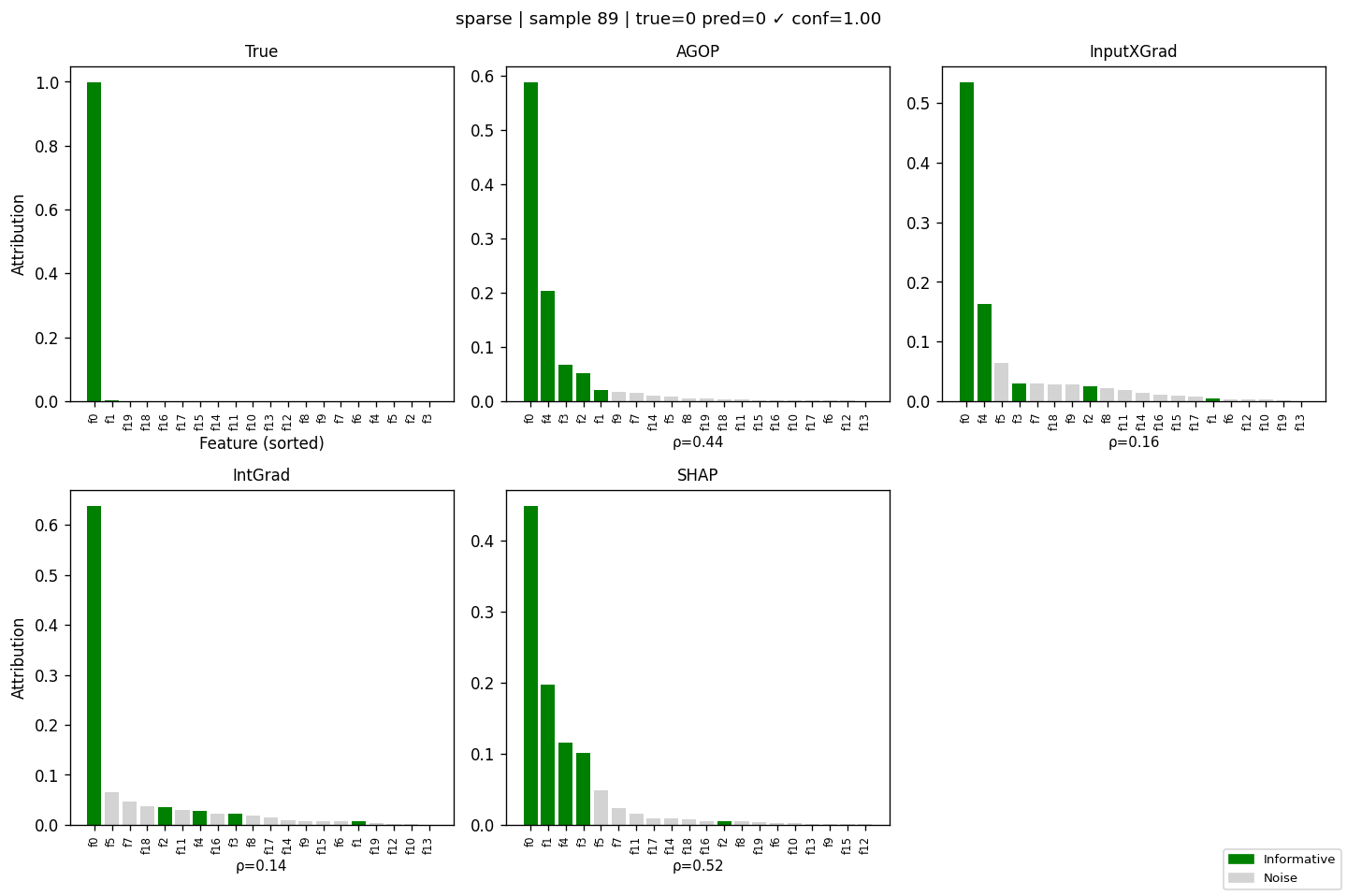}\hfill
  \includegraphics[width=0.325\linewidth]{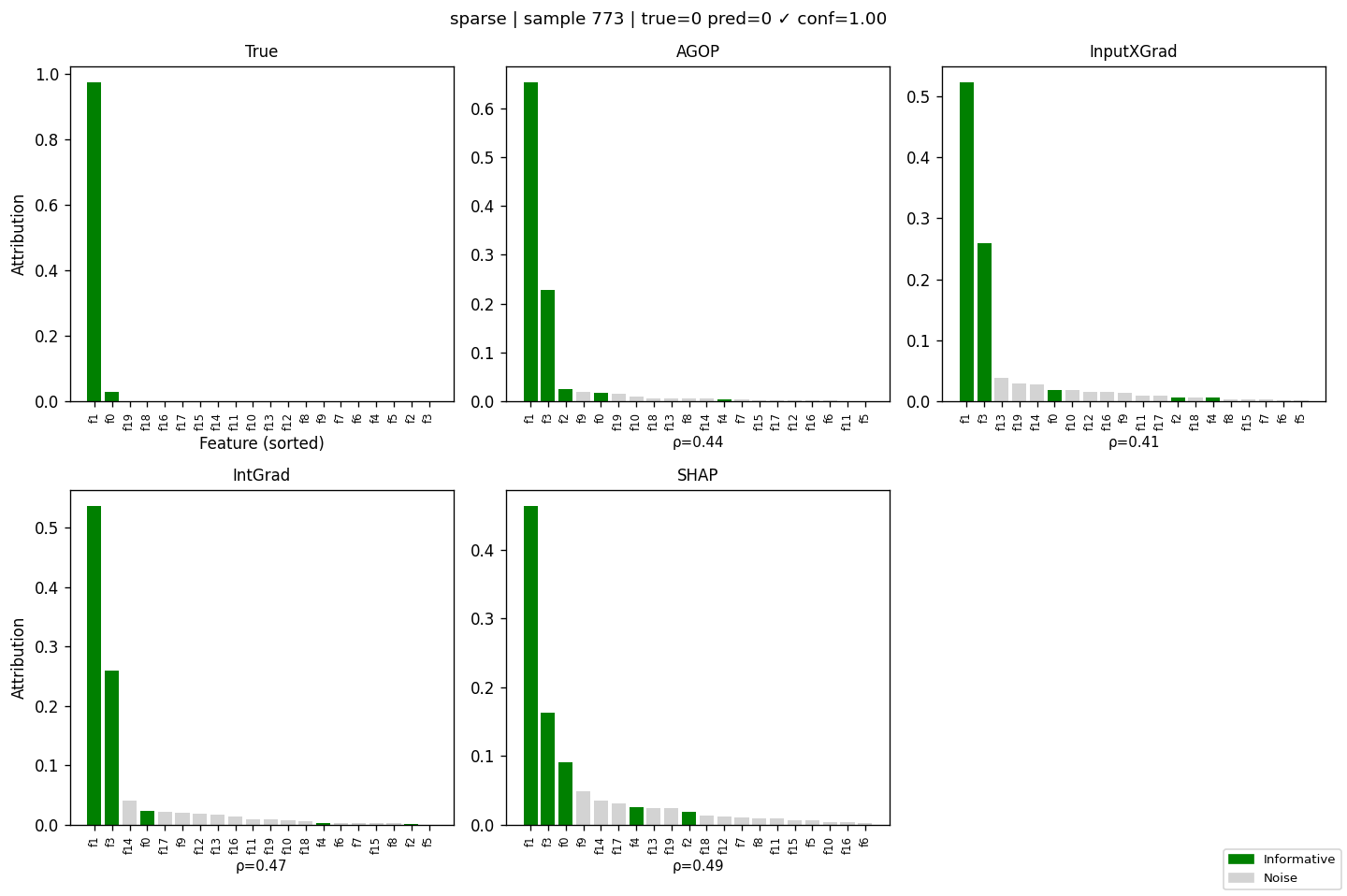}\\[2pt]
  \includegraphics[width=0.325\linewidth]{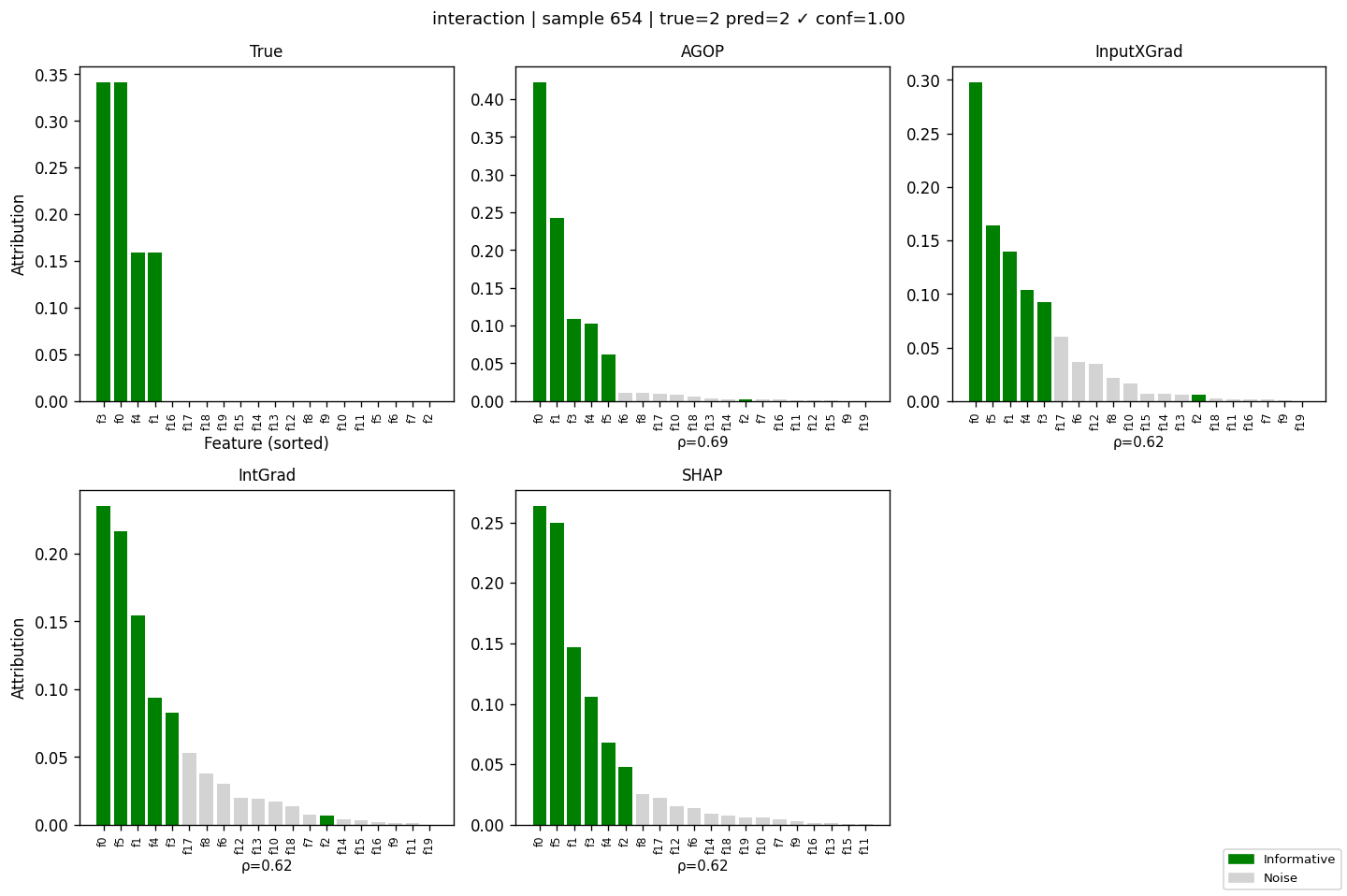}\hfill
  \includegraphics[width=0.325\linewidth]{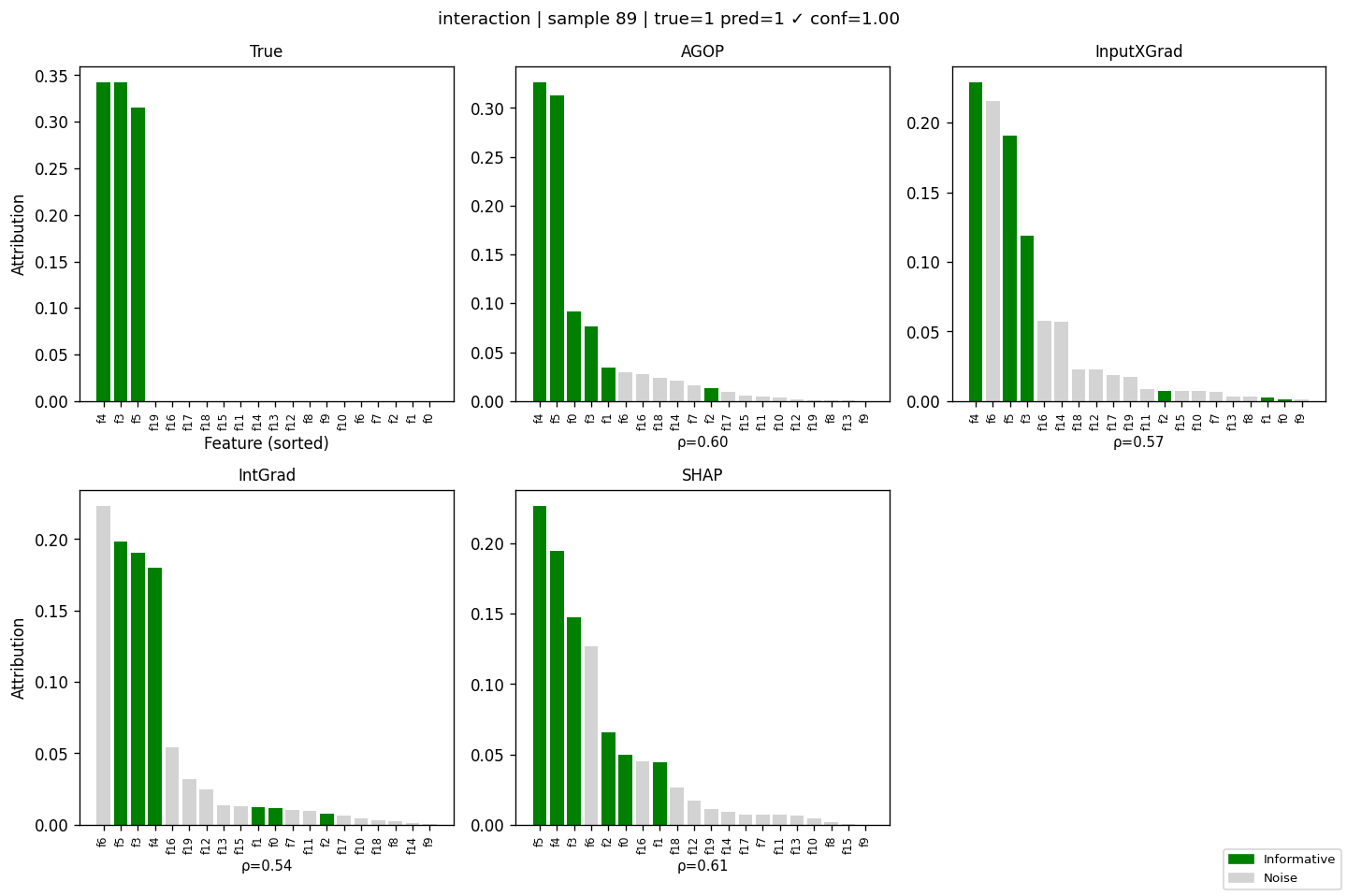}\hfill
  \includegraphics[width=0.325\linewidth]{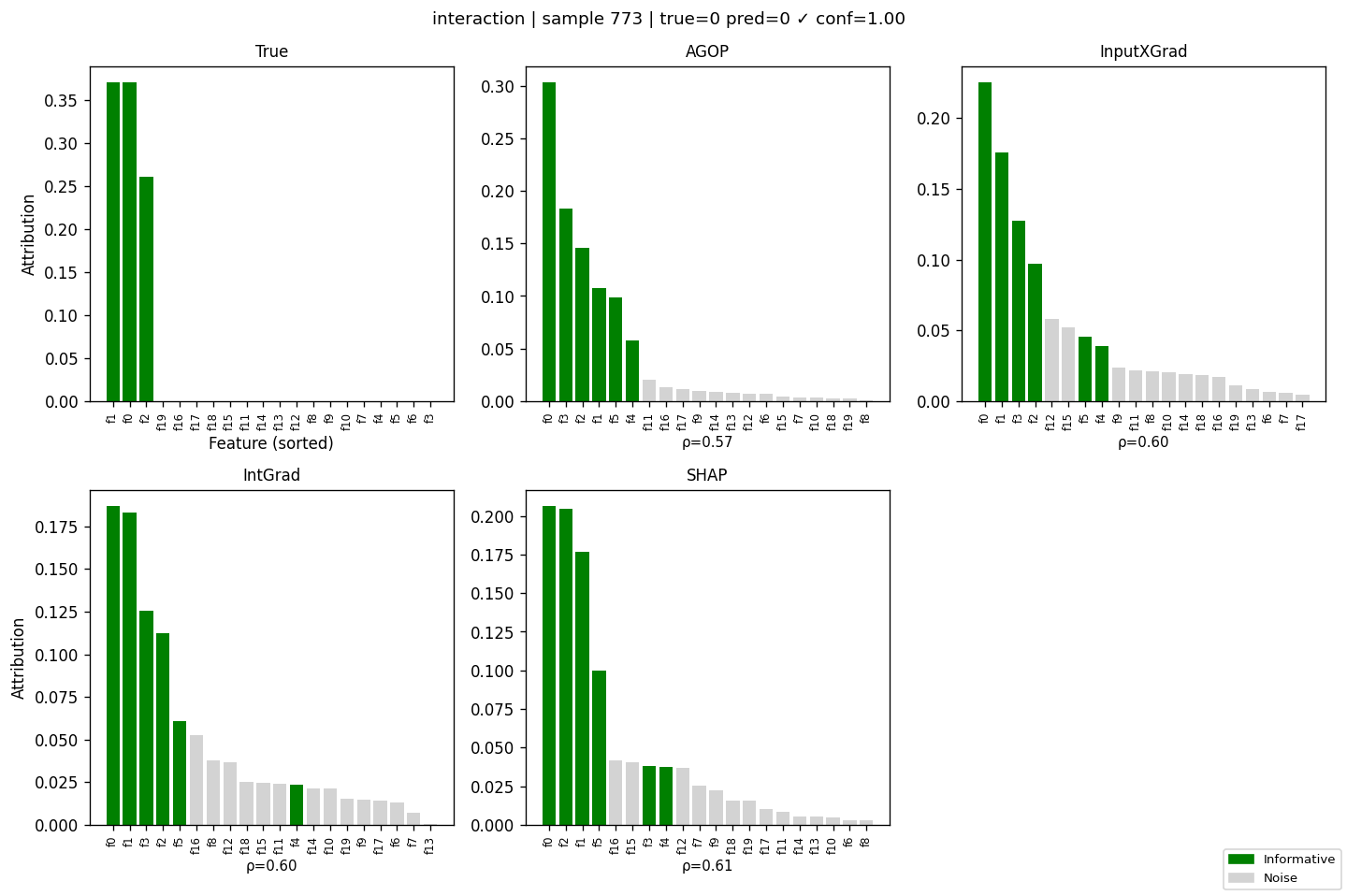}
  \caption{Per-sample attribution on the three synthetic datasets
    (rows: linear, sparse nonlinear, interaction) for three randomly
    selected test indices (columns: 654, 89, 773). Each panel's
    matplotlib title encodes the dataset, sample index, true and
    predicted class, prediction outcome ($\checkmark$/$\times$), and
    model confidence; the $x$-axis annotation of each non-True panel
    shows per-sample Spearman $\rho$ against ground truth.}
  \label{fig:per_sample}
\end{figure}

\paragraph{Per-sample qualitative analysis.}
Figure~\ref{fig:per_sample} shows attribution bar charts for three
randomly selected test indices (654, 89, 773; the same three indices
are used across all three datasets for visual comparison). Rows are
the three synthetic datasets (linear, sparse nonlinear, interaction);
columns are the three test samples. Each panel sorts features by
attribution magnitude (descending); green bars are informative
features ($f_0\ldots f_{k-1}$), grey are noise. The $x$-axis label of
each non-True panel shows the per-sample Spearman $\rho$ against
ground truth. A faithful method produces green bars dominant on the
left and grey bars on the right. LIME is omitted from the per-sample
panels for visual clarity (its results appear in the quantitative
tables); the four shown methods cover the gradient, path-integral,
game-theoretic, and second-order families.

On misclassified samples (marked $\times$ in panel titles), all methods
explain the model's \emph{wrong} prediction. High attribution to noise
features on such samples may correctly reflect what the model used to
arrive at its incorrect answer, rather than indicating method failure.

\section{Part 2: ROAR on Real Datasets}
\label{sec:roar}

\subsection{Protocol}
For each method and removal fraction $p \in \{5, 10, 20, 30, 50\}\%$:
(1) compute per-sample attributions on the training set;
(2) average across samples to obtain a global feature ranking
$\bar{r}_j = \frac{1}{n}\sum_i r_{i,j}$;
(3) replace the top-$p\%$ features with their training mean in both
train and test;
(4) retrain the model from scratch;
(5) evaluate test accuracy.
The \emph{ROAR AUC} is the area under the accuracy-vs-removal curve,
computed via trapezoidal integration over removal fractions 0--50\%;
lower AUC indicates faster degradation, i.e.,\ more faithful global
feature ranking.
All methods use the same normalized attribution averaging for the
global ranking; LIME uses 500 training samples due to per-sample cost.

\subsection{Datasets}
\textbf{Adult Income} \citep{kohavi1996scaling}: binary classification.
Raw data: 48{,}842 samples; after removing rows with missing values:
45{,}222 samples, 104 features (numeric + one-hot categorical).
80/20 stratified split $\to$ 36{,}177 train / 9{,}045 test.
Original test accuracy: \textbf{81.5\%}.

\textbf{Credit Card Default} \citep{yeh2009comparisons}: binary
classification, 30{,}000 samples, 23 features.
80/20 split $\to$ 24{,}000 train / 6{,}000 test.
Original test accuracy: \textbf{77.4\%}.

\subsection{Results}

\begin{table}[t]
\centering
\caption{ROAR AUC (lower = more faithful global feature ranking).
  Mean $\pm$ std over 4 seeds (0, 1, 2, 42). Sorted by Adult AUC.
  All pairwise differences are $\leq 0.010$; see text.}
\label{tab:roar_auc}
\small
\begin{tabular}{lcc}
\toprule
Method & Adult AUC $\downarrow$ & Credit AUC $\downarrow$ \\
\midrule
IntGrad    & $\mathbf{0.3940\pm.0011}$ & $0.3813\pm.0039$ \\
InputXGrad & $0.3942\pm.0009$          & $0.3834\pm.0014$ \\
LIME       & $0.3951\pm.0020$          & $\mathbf{0.3743\pm.0030}$ \\
SHAP       & $0.3974\pm.0023$          & $0.3755\pm.0025$ \\
AGOP-IxG   & $0.4005\pm.0016$          & $0.3787\pm.0012$ \\
\bottomrule
\end{tabular}
\end{table}

\begin{figure}[t]
  \centering
  \begin{subfigure}{0.49\linewidth}
    \includegraphics[width=\linewidth]{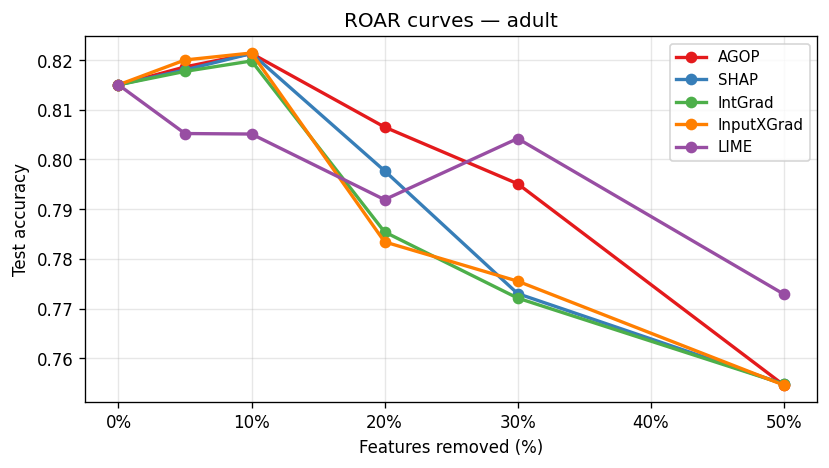}
    \caption{Adult --- ROAR curves}
  \end{subfigure}
  \begin{subfigure}{0.49\linewidth}
    \includegraphics[width=\linewidth]{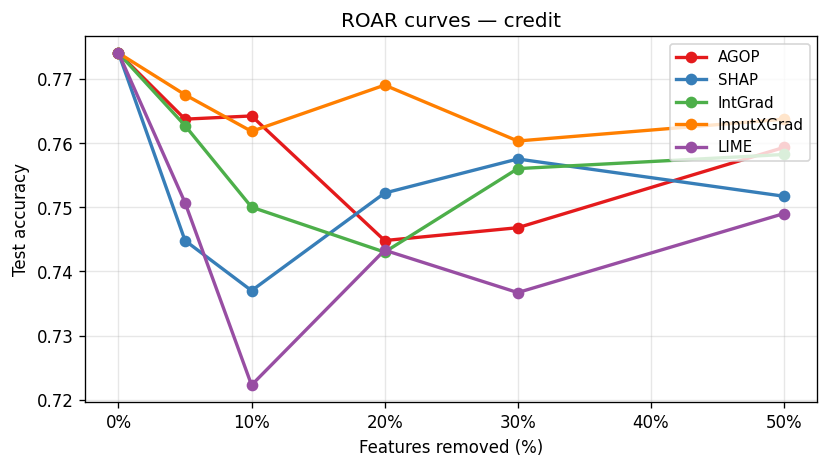}
    \caption{Credit --- ROAR curves}
  \end{subfigure}
  \begin{subfigure}{0.49\linewidth}
    \includegraphics[width=\linewidth]{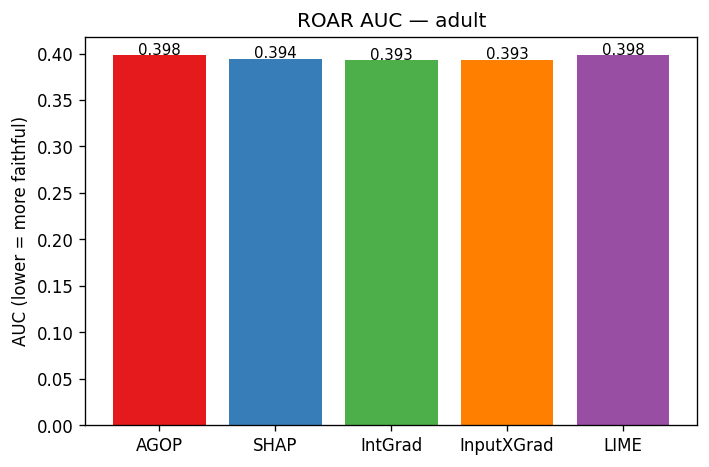}
    \caption{Adult --- ROAR AUC}
  \end{subfigure}
  \begin{subfigure}{0.49\linewidth}
    \includegraphics[width=\linewidth]{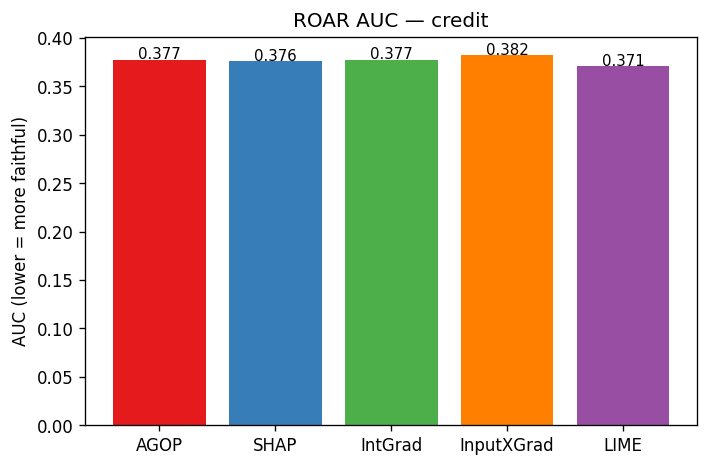}
    \caption{Credit --- ROAR AUC}
  \end{subfigure}
  \caption{ROAR results on Adult and Credit datasets.}
  \label{fig:roar}
\end{figure}

Table~\ref{tab:roar_auc} and Figure~\ref{fig:roar} show the results.
Across four seeds the AUC spread is $0.0065$ on Adult
($0.3940$--$0.4005$) and $0.0091$ on Credit ($0.3743$--$0.3834$).
Standard deviations are $\leq 0.004$, confirming that the rankings
are stable and the clustering is genuine, not seed noise.
Nevertheless, the absolute differences are small: a gap of $0.007$
on a $\approx 0.4$ baseline corresponds to $\sim 1.7\%$ relative change,
and confidence intervals of adjacent methods overlap.
The practical conclusion is that \emph{all five methods identify
comparably important features on these real datasets}; no method
dominates by a meaningful margin.

AGOP-IxG ranks last on Adult and third on Credit.
This is consistent with AGOP-IxG's design: the truncated $M_K$ filter
optimises for per-sample local attribution fidelity, while ROAR
evaluates a global feature ranking derived by averaging normalised
per-sample vectors --- a coarser and different quantity.
Crucially, AGOP-IxG's ROAR AUC is not dramatically worse; it does not
assign systematically irrelevant features as important.
The synthetic results in Part~1, which directly measure per-sample
faithfulness against known ground truth, better reflect AGOP-IxG's
intended operating regime.

On the Adult dataset, accuracy briefly \emph{exceeds} baseline for
all four gradient/Shapley methods at 5--10\% removal (AGOP-IxG,
InputXGrad, SHAP, and IntGrad all peak around 82.0--82.1\% vs.\ 81.5\%
baseline; LIME stays at or below baseline throughout). This is
consistent with correlated redundant features: removing
attributed-important features can allow the retrained model to shift
to correlated proxies.
On the Credit dataset (baseline 77.4\%), all five methods remain
\emph{below} baseline across all removal fractions; no method shows a
consistent above-baseline plateau or strictly monotone accuracy decline.

\section{Discussion: Implications for AutoML Practitioners}
\label{sec:discussion}

The combined Part 1 and Part 2 results suggest concrete guidance for
AutoML systems that emit attribution outputs:

\paragraph{Match the method to the consumer.} Different downstream
consumers of attribution have different fidelity requirements. A
per-decision explanation surface (e.g.\ a regulator-facing report on
a single denied credit application) requires local per-sample
fidelity, where AGOP-IxG performs strongly across all three synthetic
regimes. A global feature-selection step inside an AutoML pipeline
requires aggregate ranking quality, where Part 2 shows the choice
matters little among the five evaluated methods.

\paragraph{Real-time deployability.} The runtime advantage of
AGOP-IxG (Table~\ref{tab:synthetic}) translates into a deployment
gap rather than merely a benchmark number. At sub-millisecond
per-sample cost, AGOP-IxG can run \emph{inside} a model-serving
inference path and return a per-request explanation alongside every
prediction --- e.g., an adverse-action reason code attached to every
denied credit decision. SHAP at ${\sim}33$\,ms per sample is workable
for individual API calls but becomes prohibitive at scale:
explaining one million daily decisions takes ${\sim}9$\,hours of GPU
time with SHAP versus ${<}2$\,minutes with AGOP-IxG. For AutoML
pipelines that must serve explanations with every prediction, this
is the difference between real-time deployable and batch-only.

\paragraph{Ground-truth synthetic benchmarks generalize beyond
AGOP-IxG.} The three synthetic generators in this paper are intentionally
method-agnostic: any attribution method that produces a per-sample
$\mathbb{R}^d$ vector can be scored. We hope this enables future
AutoML benchmark suites to include attribution fidelity as a
first-class metric alongside accuracy and runtime.

\section{Limitations}
\label{sec:limitations}

\textbf{Synthetic datasets only for Part~1.}
Ground-truth attribution is unavailable on real data.
The synthetic tasks are deliberately structured; results on complex
real-world distributions may differ.

\textbf{Four seeds.}
Standard deviations are small and rankings are stable, but
larger-scale seed sweeps and formal significance testing
(e.g.\ Wilcoxon rank-sum) are left for future work.

\textbf{MLP only.}
All methods are evaluated on the same MLP architecture (one trained
model per Part 1 dataset; Part 2 retrains a fresh model of the same
architecture after each removal fraction). Results may differ on
tree-based or attention-based models commonly produced by AutoML
systems; extending to AutoGluon/Auto-sklearn output models is an
important direction.

\textbf{Label noise ceiling.}
The $\sigma{=}0.3$ label noise creates an $\approx$84\% Bayes-error
ceiling on sparse and interaction; linear achieves 92.1\% with the
same $\sigma$ because its score function produces larger
class-margin variance.

\textbf{LIME ROAR subsample.}
LIME's ROAR feature ranking is computed on a 500-sample subsample
due to per-sample fitting cost (other methods use the full training
set); Part~1 evaluation uses the full test set for all methods.

\section{Broader Impact}
\label{sec:impact}

Feature attribution methods are increasingly used as the
``explainability layer'' of automated decisioning systems in regulated
domains. Choosing an attribution method by reputation rather than
measured fidelity risks producing explanations that are
\emph{plausible-sounding but inaccurate} --- a failure mode with
specific downstream harms when explanations are used to contest
adverse decisions, audit fairness, or satisfy disclosure requirements.

This benchmark provides a small but rigorous evaluation surface that
practitioners and AutoML system designers can use to substantiate
method choices. The synthetic generators do not contain personal data
and the real datasets used in Part 2 (Adult, Credit) are
long-standing public benchmarks. We do not propose a deployment of
any specific method; rather, we show that no single attribution
method dominates across regimes, which should encourage AutoML
systems to expose the choice of attribution method (and its
empirical fidelity profile) to their users rather than fixing it by
convention.

\section{Conclusion}
\label{sec:conclusion}

We presented \emph{AGOP-IxG} (Section~\ref{sec:agop}) and evaluated
it against four widely-used baselines on a two-part tabular benchmark
(Sections~\ref{sec:synthetic}--\ref{sec:roar}; mean $\pm$ std over
four seeds). On the synthetic tasks AGOP-IxG performs strongly on
per-sample fidelity metrics; on real data under ROAR no method
dominates.

These results suggest that the AGOP matrix's population-level gradient
covariance, used as a per-sample gradient filter, provides a strong
inductive bias for \emph{local} per-prediction explainability without
sacrificing runtime --- a property operationally valuable inside
AutoML pipelines that must produce explanations under time budgets.
Future work should extend the evaluation to larger and more diverse
datasets, a broader set of model architectures (tree-based and
attention-based outputs from AutoML systems), and formal significance
testing across more seeds.

\bibliographystyle{apalike}
\bibliography{refs}

\appendix
\section{Implementation Details}
\label{app:impl}

\paragraph{Gradient computation.}
All gradient-based methods compute gradients w.r.t.\ the \emph{maximum
logit} (predicted class score), not the loss:
\begin{verbatim}
score_sum = model(xb).max(dim=1).values.sum()
grads = torch.autograd.grad(score_sum, xb)[0]
\end{verbatim}

\paragraph{AGOP eigenvalue truncation.}
Eigenvectors with eigenvalue $>0.01\cdot\lambda_{\max}$ are retained.
$K$ for the canonical seed-42 run: Linear~$K{=}2$, Sparse~$K{=}12$,
Interaction~$K{=}19$. $K$ varies slightly across seeds as the model
and gradient covariance structure change.

\paragraph{Reproducibility.}
Results are averaged over seeds 0, 1, 2, and 42.
For each seed, the model is trained once and saved weights are loaded
for all attribution methods and both evaluation phases.

\section{AutoML Reproducibility Checklist}
\label{app:repro_checklist}

We follow the AutoML Conference reproducibility checklist.

\paragraph{Data and reproducibility.}
\begin{itemize}
  \item Synthetic dataset generators: deterministic functions of a
    seed, fully specified in Section~\ref{sec:synthetic}.
  \item Real datasets: Adult Income \citep{kohavi1996scaling} and
    Credit Card Default \citep{yeh2009comparisons}, both publicly
    available from the UCI Machine Learning Repository.
  \item Trained model checkpoints: deterministic and reproducible
    given a seed (seeds 0, 1, 2, 42 with cuDNN deterministic mode).
\end{itemize}

\paragraph{Computational environment.}
\begin{itemize}
  \item Hardware: a single NVIDIA GPU per run (synthetic experiments
    fit in $<2$\,GB of GPU memory; ROAR retraining loops fit in
    $<4$\,GB).
  \item Software: PyTorch for model training and gradient
    computation; \texttt{shap} (DeepExplainer) and \texttt{lime}
    (\texttt{LimeTabularExplainer}) for the corresponding baselines.
  \item Seeds: 0, 1, 2, 42 for all experiments. Each seed produces a
    separate dataset, model, and attribution run.
\end{itemize}

\paragraph{Computational cost.}
\begin{itemize}
  \item Synthetic model training: $\sim$1 minute per seed per dataset
    on a single GPU.
  \item Attribution computation (per dataset, full test set):
    AGOP-IxG $\sim$0.1\,s, InputXGrad $<$0.1\,s, IntGrad $\sim$2\,s,
    SHAP $\sim$33\,s, LIME $\sim$15\,s.
  \item ROAR retraining: 5 removal fractions $\times$ 5 methods $\times$
    2 datasets $\times$ 4 seeds $=$ 200 retraining runs; $\sim$7 hours
    total on a single A100 GPU (each seed's ROAR loop is $\sim$1\,h\,45\,min;
    parallelizable across GPUs for the multi-seed runs).
\end{itemize}

\paragraph{Hyperparameter selection.}
No hyperparameter search was performed. All model and method
hyperparameters are fixed (Section~\ref{sec:methods},
Section~\ref{sec:synthetic}) and chosen from the defaults of each
reference implementation. This is intentional: the benchmark
evaluates default-configured attribution methods, which is the
operationally relevant setting for AutoML systems that cannot afford
per-deployment attribution-method tuning.

\paragraph{Reporting of variance.}
All results are reported as mean $\pm$ standard deviation over four
seeds. No method showed seed-rank instability on the synthetic
benchmarks (Section~\ref{sec:results}); ROAR rankings are stable
within $\leq 0.004$ std.

\end{document}